\documentclass[letterpaper]{article} 
\usepackage{aaai2026}  
\usepackage{times}  
\usepackage{helvet}  
\usepackage{courier}  
\usepackage[hyphens]{url}  
\usepackage{graphicx} 
\usepackage{natbib}  
\usepackage{caption} 
\usepackage{algorithm}
\usepackage{algorithmic}

\usepackage{multirow}
\usepackage{subfigure}
\usepackage{amsmath, amssymb, bm}
\usepackage{graphicx}
\usepackage{algorithm}
\usepackage{algorithmic}
\usepackage{booktabs}
\usepackage{url}
\usepackage[table]{xcolor}

\usepackage{newfloat}
\usepackage{listings}
\DeclareCaptionStyle{ruled}{labelfont=normalfont,labelsep=colon,strut=off} 
\floatstyle{ruled}
\newfloat{listing}{tb}{lst}{}
\floatname{listing}{Listing}
\title{Improving Day-Ahead Grid Carbon Intensity Forecasting by Joint Modeling of Local-Temporal and Cross-Variable Dependencies Across Different Frequencies}
\author{
    Bowen Zhang\textsuperscript{\rm 1},
    Hongda Tian\textsuperscript{\rm 1}\thanks{Corresponding Author: Hongda Tian},
    Adam Berry\textsuperscript{\rm 2},
    A. Craig Roussac\textsuperscript{\rm 3}
}
\affiliations{
    \textsuperscript{\rm 1}
    Data Science Institute, University of Technology Sydney\\
    \textsuperscript{\rm 2}
    Human Technology Institute, University of Technology Sydney\\
    \textsuperscript{\rm 3}Buildings Alive Pty Ltd.\\

    bowen.zhang@student.uts.edu.au, hongda.tian@uts.edu.au, adam.berry@uts.edu.au,
    croussac@buildingsalive.com
}

\usepackage{bibentry}

\begin{document}

\maketitle

\begin{abstract}
Accurate forecasting of the grid carbon intensity factor (CIF) is critical for enabling demand-side management and reducing emissions in modern electricity systems. Leveraging multiple interrelated time series, CIF prediction is typically formulated as a multivariate time series forecasting problem. Despite advances in deep learning-based methods, it remains challenging to capture the fine-grained local-temporal dependencies, dynamic higher-order cross-variable dependencies, and complex multi-frequency patterns for CIF forecasting. To address these issues, we propose a novel model that integrates two parallel modules: 1) one enhances the extraction of local-temporal dependencies under multi-frequency by applying multiple wavelet-based convolutional kernels to overlapping patches of varying lengths; 2) the other captures dynamic cross-variable dependencies under multi-frequency to model how inter-variable relationships evolve across the time-frequency domain. Evaluations on four representative electricity markets from Australia, featuring varying levels of renewable penetration, demonstrate that the proposed method outperforms the state-of-the-art models. An ablation study further validates the complementary benefits of the two proposed modules. Designed with built-in interpretability, the proposed model also enables better understanding of its predictive behavior, as shown in a case study where it adaptively shifts attention to relevant variables and time intervals during a disruptive event.
\end{abstract}


\section{Introduction}

\newcommand{\coo}{\ensuremath{\mathrm{CO_2}}}

Carbon dioxide (\coo) emissions from electricity generation significantly contribute to global warming \cite{F1}, with the power sector accounting for approximately 40\% of global carbon emissions~\cite{F2}. The transition to renewable energy sources is crucial for reducing emissions~\cite{F3}, yet the intermittency and uncertainty of renewables introduce challenges in maintaining a stable and low-carbon electricity supply. Indeed, variations in fuel mix and electricity demand lead to fluctuations in the carbon intensity of the grid over time.

Average carbon intensity factor (CIF) quantifies the emissions produced per kilowatt-hour (kWh) of electricity and is defined as:  
\begin{equation}
CIF_{avg,t} = \frac{\sum(E_{r,t} \times C_{r})}{\sum E_{r,t}}
\end{equation} where $\text{CIF}_{\text{avg}, t}$ denotes the average CIF across time interval $t$, $E_{r,t}$ represents the electricity generated by source type $r$ during the same period, and $C_{r}$ corresponds to the emission rate of source type $r$. 

\begin{figure}[!t]
\centering
\includegraphics[width=1\columnwidth]{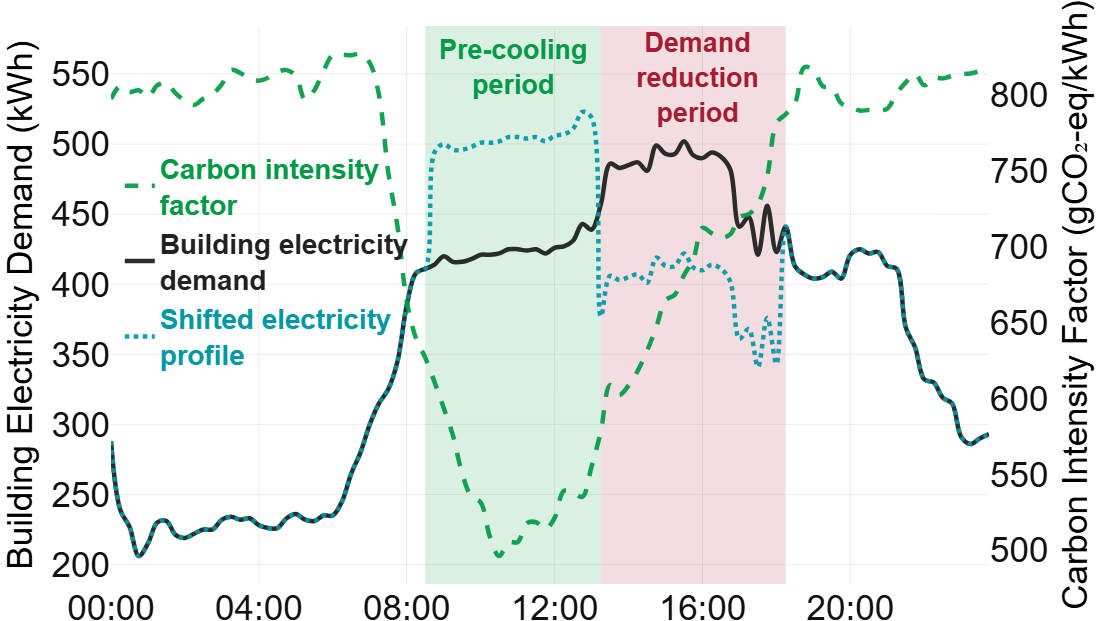}
\caption{An example from October 15, 2022, illustrates how building electricity demand can be shifted through a pre-cooling strategy during low levels of CIF periods (highlighted in green), reducing demand during high levels of CIF hours in the afternoon (highlighted in red). This helps align energy use with cleaner grid periods and contributes to emission reductions.}
\label{fig:Building_load_shifting}
\end{figure}

Accurate forecasting of the short-term grid average CIF is crucial for optimizing energy use and enabling demand-side management strategies that shift electricity consumption to periods of lower emissions \cite{F78}. This capability is beneficial for measures such as pre-cooling buildings when carbon intensity is minimal (as shown in Figure \ref{fig:Building_load_shifting}), optimizing battery storage to align with cleaner energy availability, and scheduling electric vehicle charging during low-emission periods. Enhancing this flexibility plays a key role in reducing overall \coo\ output and supporting climate targets set by, for example, the Paris Agreement~\cite{F4}.

Despite its importance, short-term CIF forecasting remains a complex challenge due to its non-linear and non-stationary nature. The inherent complexity arises from multiple interdependent factors, including energy mix variability, market dynamics, and weather-related influences on renewable energy generation (REG). Recent research has applied deep learning methods, including Artificial Neural Networks (ANNs)~\cite{F13}, Long Short-Term Memory Networks (LSTMs)~\cite{F10, F8, F9,  F11, F56}, and hybrid models~\cite{F12, F14}, which excel at identifying temporal dynamics in CIF series. However, these methods tend to exhibit reduced forecasting accuracy in volatile grid conditions, particularly in regions with high renewable energy penetration, where carbon intensity patterns have become increasingly less predictable \cite{F8}.

Most existing deep learning methods treat short-term CIF prediction as a multivariate time series (MTS) forecasting problem. A CIF dataset, often consisting of multiple interrelated time series, typically exhibits short-term temporal dynamics within each series, interactions among variables, and complex frequency characteristics. Improved modeling of this information could enhance the performance of CIF forecasting. As shown below, existing studies still have challenges in modeling local-temporal dependencies (LTD), cross-variable dependencies (CVD), and multi-frequency information (MFI). Based on this insight, we propose targeted solutions to address each of these issues accordingly.

\paragraph{Local-Temporal Dependencies} LTD involves short-term periods and the temporal patterns in adjacent time steps within each individual input time series. For instance, in a CIF series at hourly intervals, LTD may represent local CIF patterns that emerge within a few hours. Existing segment-based approaches~\cite{F20, F15, F21, F83, F86} attempt to capture LTD using fixed-length, non-overlapping segments but often face challenges in capturing dependencies across adjacent segments and extracting fine-grained temporal patterns within each segment~\cite{F91}. To address these challenges, we adopt an adaptable segmentation strategy~\cite{F90} that divides the time series into overlapping patches with varying lengths. This design enables the model to capture local patterns at diverse and fine-grained temporal resolutions. Furthermore, continuous wavelet transform (CWT) functions are embedded as convolutional kernels to extract localized time-frequency features.

\paragraph{Cross-Variable Dependencies}  
Different variables in CIF forecasting often interact with each other, revealing interlinked dynamics that influence CIF. For instance, REG, grid load demand (GLD), and non-renewable energy generation (NEG) jointly affect CIF. An increase in REG may reduce CIF, but if GLD rises simultaneously and is met by NEG, the overall CIF may still increase. These dependencies are also dynamic and can shift under varying grid conditions. Current methods~\cite{F21, F15, F82, F87} attempt to model these sorts of dependencies but only capture partial relationships among variables, often assuming static or linear interactions, which limits their ability to fully capture complex and evolving dependencies. To address these limitations, we adopt Local Multiple Regression (LMR)~\cite{F23, F24} to dynamically estimate time-varying dependencies among input variables. Moreover, all the possible combinations of the input variables are systematically encoded into structured tensors via LMR, enabling Convolutional Neural Networks (CNNs) to jointly process and learn expressive representations of CVD.

\paragraph{Multi-Frequency Information}
CIF dataset inherently reflects multi-frequency variation patterns, including both high-frequency fluctuations and low-frequency trends shaped by diverse grid dynamics. A typical example of a higher-frequency pattern is the ramping up of alternative (often higher-emission) generation sources in response to sudden declines in solar power output caused by transient weather events. Conversely, a notably lower-frequency pattern within the same context is the seasonal variation in average solar irradiance, which directly influences the average output of solar power systems. Existing methods~\cite{F76, F66, F84} typically employ a single wavelet basis, which restricts the ability to extract diverse and complementary frequency characteristics from MTS data. Additionally, no existing research considers modeling CVD under multi-frequency, limiting their ability to capture the frequency-aware dynamic CVD. To handle these issues, multiple types of wavelet functions are employed to leverage their distinct sensitivities to different signal structures. Additionally, the CWT is integrated into the LMR framework to capture CVD at different time-frequency resolutions.

In summary, the key contributions of this work are outlined as follows: 

\begin{itemize}

  \item A Local-Temporal Multi-Wavelet Kernel Convolution (LT-MWKC) module is proposed to extract localized temporal patterns across multiple frequencies by combining adaptive segmentation with diverse wavelet-based convolutional kernels.
  
  \item We propose a Cross-Variable Dynamic-Wavelet Correlation Convolution (CV-DWCC) module to capture dynamic CVD under multi-frequency. It models how cross-variable relationships evolve across the time-frequency domain through capturing comprehensive interaction patterns among input variables.

  \item By integrating the two proposed modules, our model achieves state-of-the-art (SOTA) predictive accuracy in short-term CIF forecasting, outperforming contemporary methods across four Australian electricity markets characterized by volatile and steady grid CIF patterns. Moreover, the model is designed with built-in interpretability, enabling an improved understanding of its predictive behavior.

\end{itemize}

\section{Proposed Method} \label{Methods}

In MTS forecasting, given historical observations \(\mathbf{X} = \{\mathbf{x}_1, \ldots , \mathbf{x}_T\} \in \mathbb{R}^{T \times N}\) consisting of \(T\) time steps and \(N\) variables, the objective typically involves forecasting future observations \(\mathbf{Y} = \{\mathbf{x}_{T+1}, \ldots , \mathbf{x}_{T+S}\}\ \in \mathbb{R}^{S \times N}\), where \(S\) denotes the forecast horizon. In this study, which focuses solely on predicting CIF, the objective simplifies to forecasting the next \(S\) time steps \(\mathbf{e} = \{e_{T+1}, \ldots, e_{T+S}\} \in \mathbb{R}^{S}\).

\begin{figure}[!t]
\centering
\includegraphics[width=1\columnwidth]{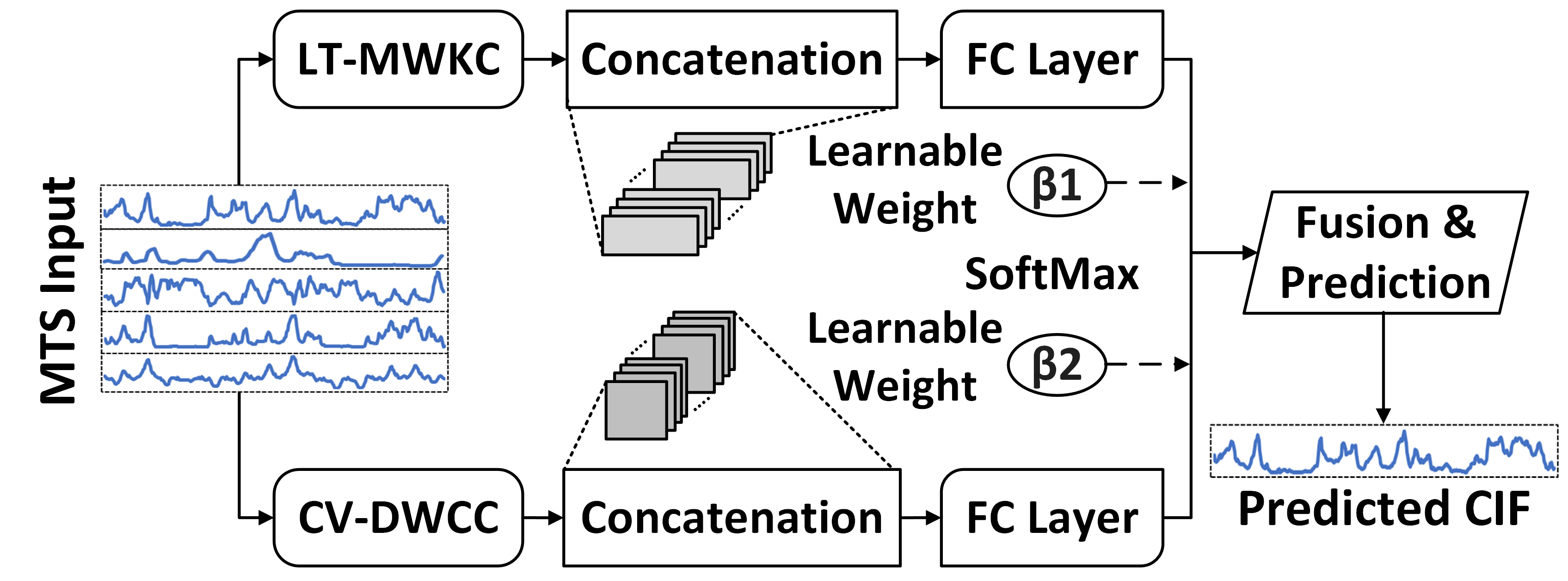}
\caption{Overall architecture of the proposed model.}
\label{fig:All_Structure_Sim}
\end{figure}

The proposed forecasting model, illustrated in Figure \ref{fig:All_Structure_Sim}, integrates two parallel modules: (1) LT-MWKC, which captures LTD enriched by MFI, and (2) CV-DWCC, which models CVD under multi-frequency. The outputs from each module are first concatenated and passed through fully connected (FC) layers, followed by a softmax-weighted fusion mechanism that adaptively integrates their contributions based on learned importance scores. 

\subsection{Local-Temporal Multi-Wavelet Kernel Convolution Module}\label{LT_MWKC}

\begin{figure}[!t]
\centering
\includegraphics[width=\columnwidth]{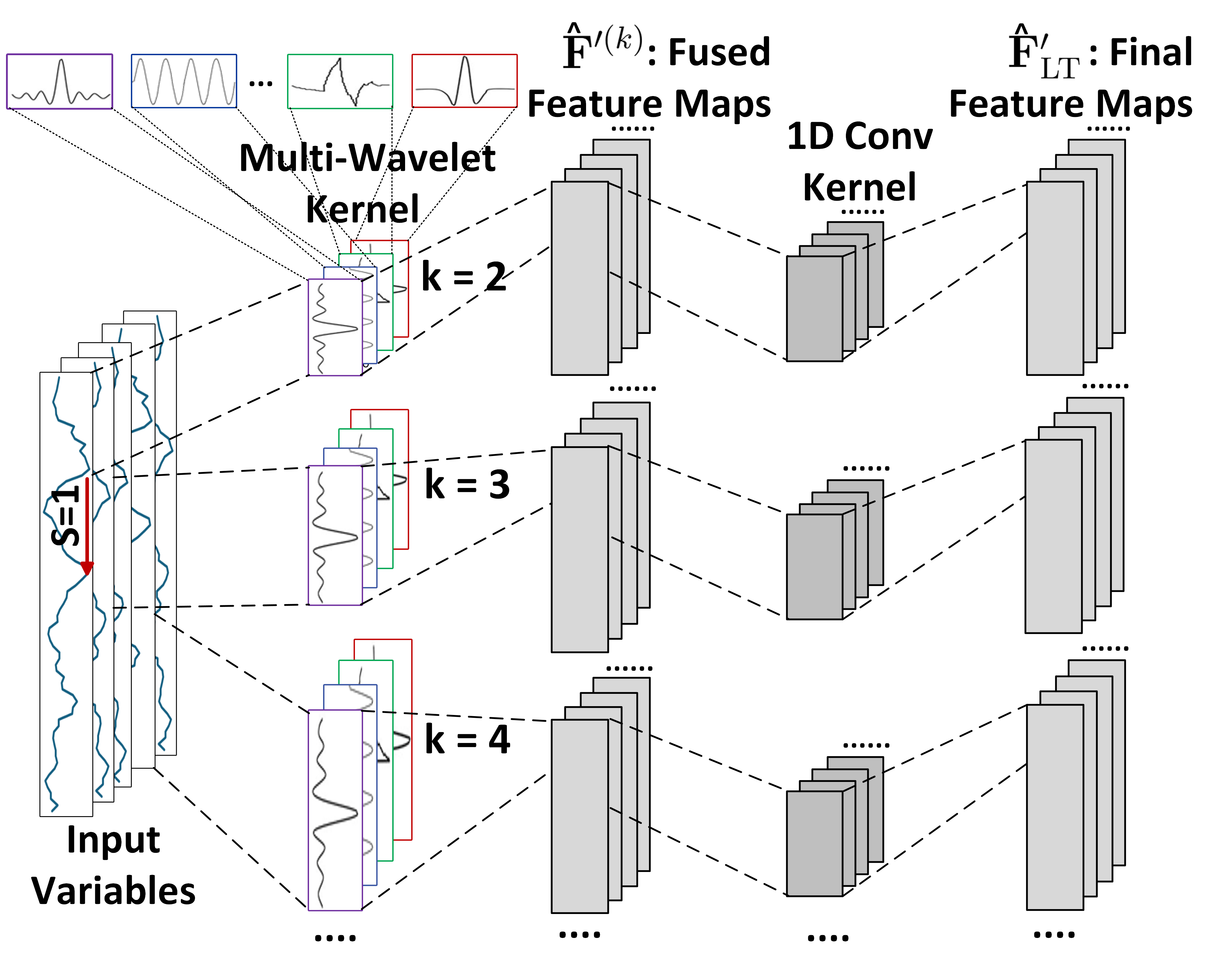}
\caption{The architecture of LT-MWKC.}
\label{fig:LT-MWKC}
\end{figure}

To effectively capture LTD in CIF forecasting, the proposed LT-MWKC module segments the input into overlapping patches with varied lengths, allowing for modeling of LTD at varying time spans, as shown in Figure~\ref{fig:LT-MWKC}. This varying-length design enables the model to capture patterns and trends that may span over different local-temporal durations. The overlapping design ensures continuity across adjacent time periods, preserving fine-grained transitions and reducing boundary effects. Each patch is processed by multi-wavelet in parallel kernels to extract a wealth of wavelet-specific MFI, while wavelet-based 1D convolutions capture LTD across varying temporal durations.

Specifically, CNNs have proven effective in time series analysis~\cite{F33,F36,F20} due to their ability to capture hierarchical temporal features. Moreover, CNNs can be structured with filters of varying sizes, allowing them to capture patterns at multiple temporal resolutions~\cite{F79, F80}. As shown in Figure~\ref{fig:LT-MWKC}, 1D convolutional filters with varying kernel sizes are applied to the MTS input using a stride of one. Larger kernels enable the model to aggregate information over longer time intervals, facilitating the extraction of extended LTD. Furthermore, the CWT offers powerful multi-frequency analysis capabilities for non-stationary signals~\cite{F26, F25}, due to its continuous translation and scale-sensitive structure, which enables effective time-frequency feature extraction as an alternative to standard 1D convolutions. Instead of fixed-shape kernels, we adopt multiple types of wavelet functions (e.g., Morlet, Mexican Hat) as convolutional filters, leveraging their complementary time-frequency characteristics to extract a wealth of wavelet-specific MFI. 

Consequently, the convolutional process is expected to yield feature maps of various dimensions. Given the transposed input MTS data $\mathbf{X}^{tr} \in \mathbb{R}^{N \times T}$, the multi-wavelet convolutional operation can be formulated as:
\begin{equation}
\label{MWKC_Operation}
\begin{split}
\mathbf{F}^{(k)} &= \text{MWKC1D}\big(
    \mathbf{X}^{tr},
    \{\boldsymbol{\Psi}_m^{(k)}\}_{m=1}^{M},
    \text{stride}=1
) \\
&\in \mathbb{R}^{(N \times Z_k) \times (T-k+1)}.
\end{split}
\end{equation} where $N$ denotes the number of input variables and $T$ is the number of time steps. $Z_k$ is the number of filters corresponding to kernel size $k \in \{2, 3, \ldots, d\}$. $\boldsymbol{\Psi}_m^{(k)} \in \mathbb{R}^{(N \times Z_k) \times k}$ represents the $m$-th wavelet kernel of length $k$, derived from a distinct wavelet function (e.g., Morlet, Mexican Hat), and $M$ is the total number of wavelet types used. The operator $\text{MWKC1D}(\cdot)$ denotes the 1D convolution operation using multi-wavelet kernels with stride 1 along the temporal axis.

To integrate the information captured by different wavelet types, the outputs of the convolutional branches are fused through a learnable weighted mechanism. The fused output is computed as:
\begin{equation}
\label{MWKC_Fusion}
\mathbf{F}'^{(k)} = \sum_{m=1}^{M} \alpha_m \cdot (\boldsymbol{\Psi}_m^{(k)} * \mathbf{X}^{tr}),
\end{equation}
where $(\boldsymbol{\Psi}_m^{(k)} * \mathbf{X}^{tr})$ denotes the convolution of the input with the $m$-th wavelet kernel, and $\alpha_m$ is a learnable scalar that adjusts the contribution of each wavelet kernel. Finally, additional 1D convolutional blocks enhance the extracted representations $\mathbf{F}'^{(k)}$ by deepening local-temporal feature extraction and generating the final feature maps \( \mathbf{F}'_{\text{LT}}\).

\subsection{Cross-Variable Dynamic-Wavelet Correlation Convolution Module}\label{CF_DWCC}

\begin{figure*}[!t]
\centering
\includegraphics[width=0.95\textwidth]{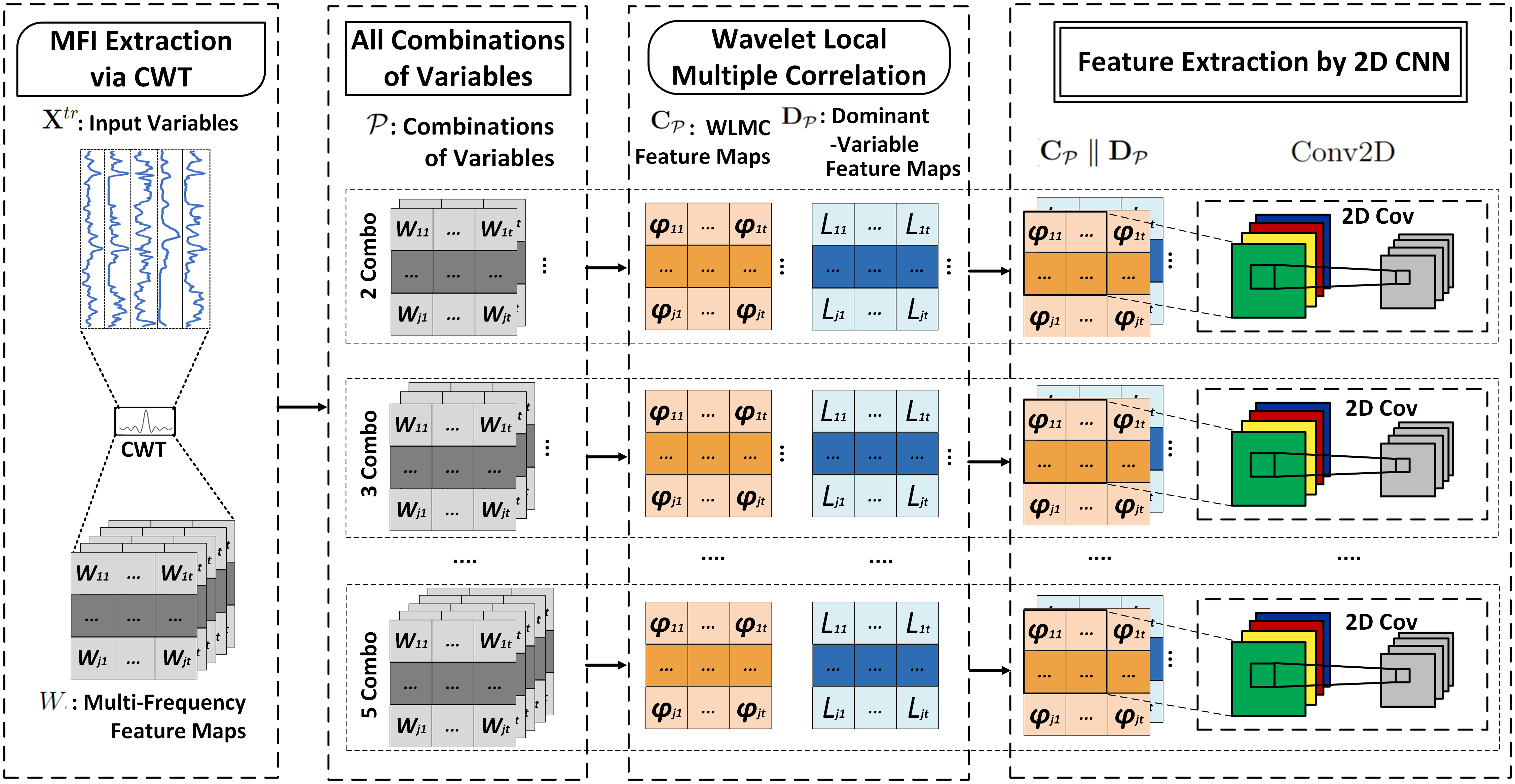}
\caption{The architecture of CV-DWCC.}
\label{fig:CF_MWCC}
\end{figure*}

To effectively capture CVD under multi-frequency, we propose the CV-DWCC module, which systematically quantifies dynamic correlations among all the possible combinations of the input variables at multiple frequency resolutions, as illustrated in Figure~\ref{fig:CF_MWCC}. The module first computes wavelet local multiple correlation (WLMC) over time and frequency for each variable combination, capturing how their interactions evolve at different time-frequency resolutions. The dominant variable that maximizes the WLMC is selected within each combination, allowing the model to focus on the most influential variable driving the interaction and reducing noise from less relevant contributors. By learning rich correlation representations through 2D convolution, the CV-DWCC module enables the model to effectively capture complex and evolving inter-variable structures.

WLMC~\cite{F23, F24} is a wavelet-based method for measuring time-evolving, multi-frequency correlations in non-stationary MTS. It identifies dominant variables by computing the maximum coefficient of determination from locally weighted regressions on wavelet coefficients, making it well-suited for capturing dynamic CVD in CIF forecasting. The WLMC builds upon the concept of wavelet-based LMR, as introduced in \cite{F23,F24}. In our study, it is applied to the transposed input MTS data $\mathbf{X}^{tr} \in \mathbb{R}^{N \times T}$. For each target variable $\mathbf{X}_i^{tr} \in \mathbf{X}^{tr}$ and a fixed time point $s \in \{1, \dots, T\}$, the LMR loss is estimated as the weighted sum of squared residuals:
\begin{equation}
L_s = \sum_{t} \theta(t - s) \left[f_s\left(X_{-i,t}\right) - X_{i,t}\right]^2,
\label{eq:local_regression_loss}
\end{equation}
where $\theta(t - s)$ is a temporal weighting function assigning higher importance to observations near $s$, and $f_s(X_{-i,t})$ denotes the local regression function estimated from all variables excluding the target $\mathbf{X}_i$. The corresponding local coefficient of determination is given by:
\begin{equation}
R^2_s = 1 - \frac{\sum_{t}\theta(t - s) \left[f_s(X_{-i,t}) - X_{i,t}\right]^2}{\sum_{t}\theta(t - s)\left[X_{i,t}-\bar{X}_{i,s}\right]^2},
\label{eq:local_R2}
\end{equation}
where $\bar{X}_{i,s}$ is the locally weighted mean of $\mathbf{X}_i$.

To capture time-frequency localized CVD, we apply CWT instead of Maximal Overlap Discrete Wavelet Transform (MODWT) used in \cite{F23, F24} to each input variable. CWT provides better time-frequency resolution and is particularly effective for modeling short-term variability and MFI. Let $\mathbf{W}_{j,t} = (w_{1,j,t}, \dots, w_{N,j,t})$ represent the CWT coefficients of each variable at scale $j = 1, \dots, J$. The WLMC coefficient $\varphi_{X,s}(j)$ at each scale $j$ and time $s$ is computed as the square root of the local coefficient of determination corresponding to the dominant variable:
\begin{equation}
\varphi_{X,s}(j) = \sqrt{R^2_{j,s}(i^*_{j,s})}, \text{with } i^*_{j,s} = \mathop{\mathrm{arg\,max}}_{i \in \{1,\dots,N\}} R^2_{j,s}(i),
\label{eq:WLMC_R2}
\end{equation}
where $R^2_{j,s}(i)$ measures the local regression fit for variable $\mathbf{X}_i$ at time $s$ and scale $j$, and $i^*_{j,s}$ denotes the index of the dominant variable that maximizes $R^2$ at that time and scale.

WLMC is applied across all the possible combinations of the input variables derived from the input  $\mathbf{X}^{tr}$, forming structured correlation tensors $\mathbf{C} \in \mathbb{R}^{\mathcal{N} \times J \times T}$, where $\mathcal{N}$ is the number of combinations, $J$ is the number of CWT scales, and $T$ is the time dimension. In parallel, the dominant-variable for each variable combination is selected as the one that maximizes the local multiple correlation $R^2_{j,s}(i)$ at each scale $j$ and time step $s$. These WLMC feature maps and corresponding dominant-variable feature maps are then processed jointly by 2D convolutional layers, formulated as follows:

\begin{equation}
\mathbf{F}_{\mathcal{P}} = \mathrm{Conv2D}([\mathbf{C}_{\mathcal{P}} \parallel \mathbf{D}_{\mathcal{P}}], \mathbf{K}_{\text{CV}}),
\label{eq:cf_dwcc}
\end{equation}
where $\mathbf{C}_{\mathcal{P}} \in \mathbb{R}^{J \times T}$ is the WLMC feature map for a given variable combination $\mathcal{P}$, $\mathbf{D}_{\mathcal{P}} \in \mathbb{R}^{J \times T}$ is the corresponding dominant-variable feature map, $\parallel$ denotes channel-wise concatenation, and $\mathbf{K}_{\text{CV}}$ is a learnable 2D convolutional kernel.

\begin{table*}[!ht]
\centering
\setlength{\tabcolsep}{2mm} 
\footnotesize 
\begin{tabular}{@{}l
>{\centering\arraybackslash}p{0.9cm}>{\centering\arraybackslash}p{0.9cm}>{\centering\arraybackslash}p{0.9cm}
>{\centering\arraybackslash}p{0.9cm}>{\centering\arraybackslash}p{0.9cm}>{\centering\arraybackslash}p{0.9cm}
>{\centering\arraybackslash}p{0.9cm}>{\centering\arraybackslash}p{0.9cm}>{\centering\arraybackslash}p{0.9cm}
>{\centering\arraybackslash}p{0.9cm}>{\centering\arraybackslash}p{0.9cm}>{\centering\arraybackslash}p{0.9cm}@{}}
\toprule
\multirow{3}{*}{\textbf{Model}} 
& \multicolumn{3}{c}{\textbf{NSW}} 
& \multicolumn{3}{c}{\textbf{SA}} 
& \multicolumn{3}{c}{\textbf{QLD}} 
& \multicolumn{3}{c}{\textbf{VIC}} \\
\cmidrule(lr){2-4} \cmidrule(lr){5-7} \cmidrule(lr){8-10} \cmidrule(lr){11-13}
& \scriptsize RMSE & \scriptsize MAE & \scriptsize SMAPE
& \scriptsize RMSE & \scriptsize MAE & \scriptsize SMAPE
& \scriptsize RMSE & \scriptsize MAE & \scriptsize SMAPE
& \scriptsize RMSE & \scriptsize MAE & \scriptsize SMAPE \\
\midrule
Proposed model & \textbf{47.09} & \textbf{36.56} & \textbf{5.25} 
               & \textbf{99.27} & \textbf{70.29} & \textbf{32.21} 
               & \textbf{31.17} & \textbf{19.46} & \textbf{2.96} 
               & \textbf{79.36} & \textbf{57.93} & \textbf{7.13} \\
LSTM & 55.57 & 43.59 & 6.25 
     & 147.51 & 109.07 & 47.19 
     & 50.01 & 32.28 & 5.02 
     & 114.06 & 84.31 & 10.40 \\
SVR & 67.58 & 46.76 & 6.73 
    & 154.75 & 116.42 & 53.29 
    & 58.24 & 39.23 & 6.01 
    & 145.87 & 110.35 & 13.55 \\
LSTNet & 50.53 & 39.45 & 5.68 
       & 125.35 & 96.23 & 43.37 
       & 52.08 & 38.98 & 5.99 
       & 98.85 & 76.72 & 9.53 \\
Crossformer & 51.68 & 40.72 & 5.82 
            & 130.51 & 101.15 & 45.01 
            & 39.81 & 28.71 & 4.47 
            & 91.08 & 70.88 & 8.78 \\
Informer & 66.80 & 50.10 & 7.35 
         & 158.95 & 124.54 & 54.30 
         & 51.91 & 37.42 & 5.82 
         & 141.54 & 112.64 & 13.81 \\
TimesNet & 49.33 & 38.20 & 5.49 
         & 132.98 & 98.21 & 43.61 
         & 38.41 & 25.61 & 3.87 
         & 96.11 & 72.28 & 8.80 \\
DLinear & 50.57 & 39.26 & 5.63 
        & 135.74 & 106.64 & 47.03 
        & 42.50 & 28.85 & 4.54 
        & 97.67 & 74.79 & 9.31 \\
NonStaFormer & 49.12 & 38.17 & 5.51 
             & 126.77 & 96.07 & 43.10 
             & 39.62 & 26.82 & 4.04 
             & 95.09 & 73.36 & 8.89 \\
PatchTST & 48.79 & 37.45 & 5.38 
         & 139.49 & 102.92 & 44.71 
         & 39.08 & 25.64 & 3.87 
         & 99.80 & 74.75 & 9.10 \\
iTransformer & 48.08 & 37.10 & 5.33 
             & 132.99 & 98.02 & 43.32 
             & 48.20 & 33.36 & 4.99 
             & 110.66 & 85.58 & 10.37 \\
TimeMixer & 47.79 & 36.79 & 5.28 
          & 134.50 & 100.95 & 44.67 
          & 39.08 & 25.99 & 3.93 
          & 96.31 & 72.34 & 8.81 \\
WPMixer & 49.64 & 38.19 & 5.47 
        & 143.20 & 106.69 & 46.55 
        & 39.02 & 25.51 & 3.84 
        & 102.08 & 76.96 & 9.37 \\
\bottomrule
\end{tabular}
\captionsetup{justification=raggedright,singlelinecheck=false}
\caption{Overall performance of the proposed model and SOTA models based on RMSE (g CO\textsubscript{2}-e/kWh), MAE (g CO\textsubscript{2}-e/kWh), and SMAPE (\%) in NSW, SA, QLD, and VIC.}
\label{model_comparison}
\captionsetup{justification=centering} 
\end{table*}

\section{Experiment}\label{Experiment}
\subsection{Data Sets}

To assess the performance of the proposed model for CIF forecasting, this paper has selected four states in Australia: New South Wales (NSW), South Australia (SA), Queensland (QLD), and Victoria (VIC). The selection of these regions for our case study aims to validate the ability of the proposed model to handle the intricacies of CIF forecasting in conventional and renewable-centric energy systems. 

NSW, in southeastern Australia, is the most populous state and relies primarily on coal and gas for its electricity. VIC, also in the southeast, shows similar non-renewable reliance but with a higher share from solar and hydro and less from wind. QLD, in the northeast, remains strongly fossil-fuel-based, with coal and gas as the dominant sources. In contrast, SA, in the south-central region, is renewable-dominated, with wind and solar as the main sources. We note that these preliminary findings highlight the potential of the proposed model, and emphasize that further evaluation on a broader test set is needed to confirm its generalizability. The datasets span from January 1, 2020, to December 31, 2023, with hourly intervals.

Five relevant input variables, including CIF, GLD, REG, NEG, and temperature, are considered, which have been commonly used in prior studies~\cite{F8, F11, F12, F14}. These datasets are sourced from the Australian Energy Market Operator (AEMO)~\cite{F68} and Open Platform for National Electricity Market (OpenNEM)~\cite{F69} platforms. Rather than aiming for optimal CIF forecasting through an exhaustive dataset, this study highlights promising forecasting performance achieved by the model using a limited set of input variables.

\subsection{Experimental Setup}
The model performance is evaluated using three metrics: Root Mean Squared Error (RMSE), Mean Absolute Error (MAE), and Symmetric Mean Absolute Percentage Error (SMAPE).

To ensure a comparative analysis and validate the effectiveness of our proposed model, we select benchmark models commonly used in CIF forecasting and SOTA methods in MTS forecasting. We include LSTM and Support Vector Regression (SVR), two widely used models in CIF forecasting research, for performance comparison~\cite{F10, F8, F9, F14, F11}. Additionally, we incorporate SOTA deep learning methods designed for MTS forecasting, including LSTNet~\cite{F62}, Crossformer~\cite{F21}, Informer~\cite{F17}, TimesNet~\cite{F20}, DLinear~\cite{F32},  Non-stationary Transformers (NonStaFormer)~\cite{F65}, PatchTST~\cite{F19}, iTransformer~\cite{F87}, TimeMixer~\cite{F88}, and WPMixer~\cite{F86}.

The dataset is structured into samples, each consisting of 24 hours of input variables and the subsequent 24 hours of CIF as prediction targets. These input-output pairs are generated using a sliding window method with a 1-hour step, producing a total of 35,017 samples per state. A 5-fold cross-validation strategy is employed, with final results averaged over all five test folds. Grid search is used to individually tune all models, including our proposed model, for each state to ensure fair comparison in the CIF forecasting task.

\subsection{Overall Results}

The performance of all models was assessed using RMSE, MAE, and SMAPE for NSW, SA, QLD, and VIC, as detailed in Table \ref{model_comparison}. Across all evaluation metrics, the proposed model consistently achieved the most accurate predictions in four regions.

\begin{figure*}[htbp]
\centering

\begin{minipage}{\columnwidth}
    \centering
    \includegraphics[width=\columnwidth]{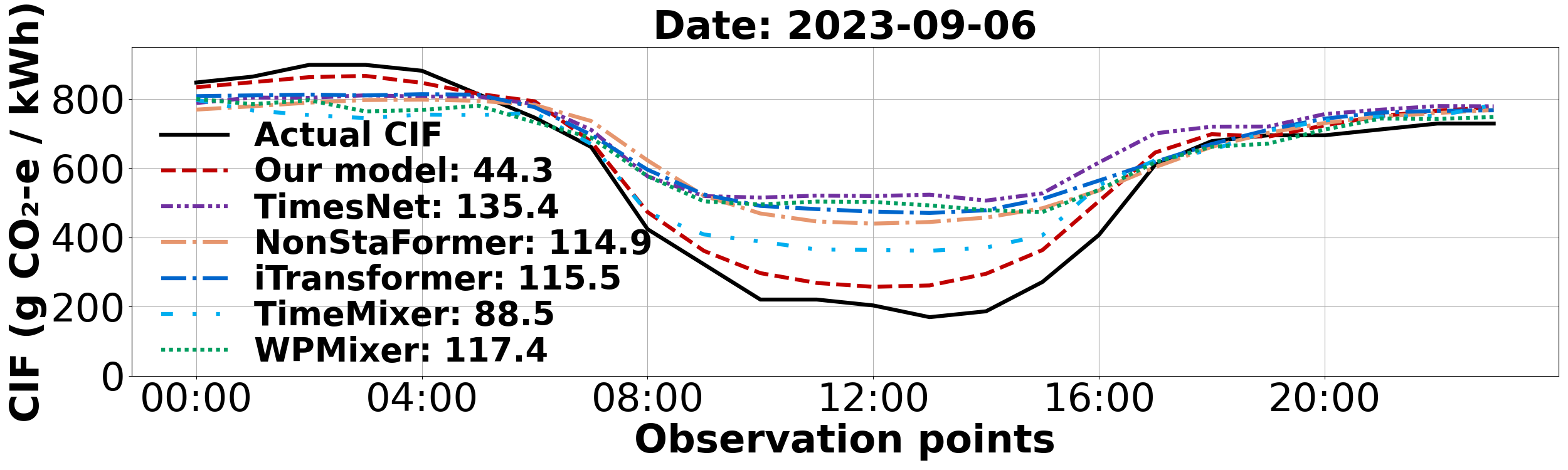}
    \textbf{(a)}
\end{minipage}
\begin{minipage}{\columnwidth}
    \centering
    \includegraphics[width=\columnwidth]{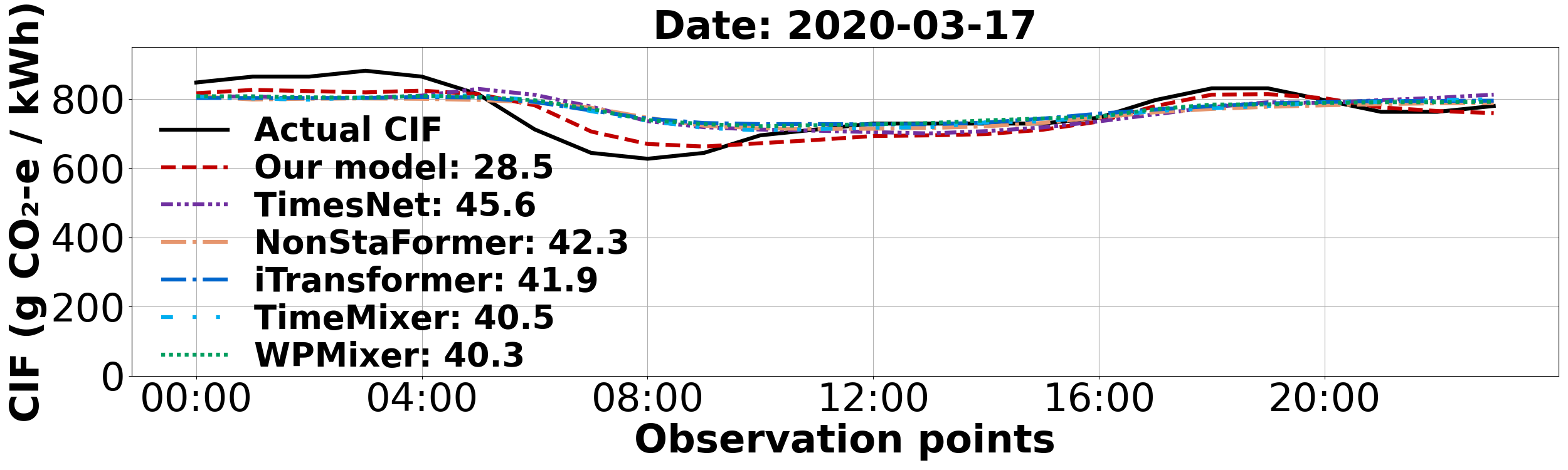}
    \textbf{(b)}
\end{minipage}

\begin{minipage}{\columnwidth}
    \centering
    \includegraphics[width=\columnwidth]{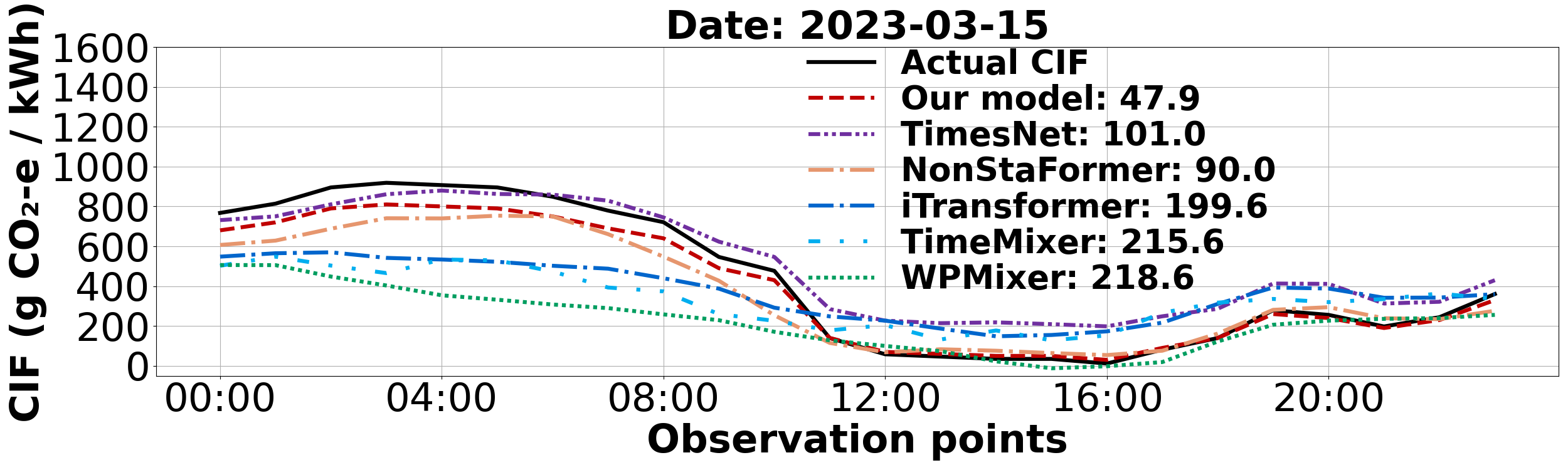}
    \textbf{(c)}
\end{minipage}
\begin{minipage}{\columnwidth}
    \centering
    \includegraphics[width=\columnwidth]{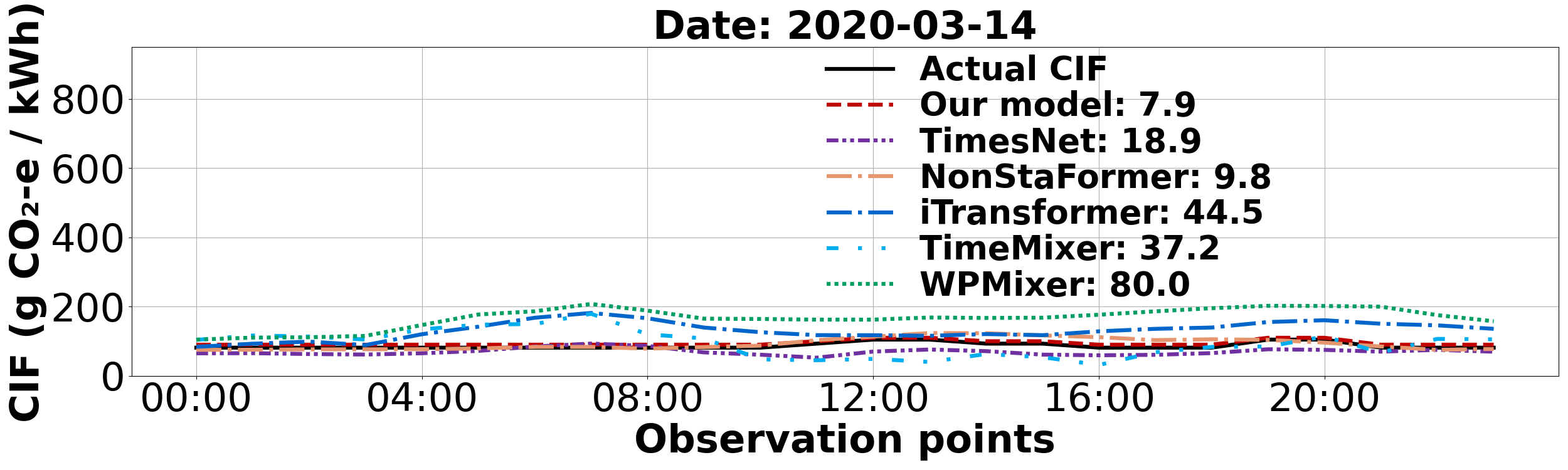}
    \textbf{(d)}
\end{minipage}

\caption{Forecasting results with MAE in four example cases: CIF with the largest variation in NSW (a) and SA (c); CIF with the smallest variation in NSW (b) and SA (d).}
\label{fig:ape_cases}
\end{figure*}

We illustrate the forecasting performance of our proposed model against SOTA models under varying grid conditions by presenting results on four representative days in Figure \ref{fig:ape_cases}. These cases are selected based on the largest and smallest variations in the CIF observed in NSW and SA over the test period. In Figures \ref{fig:ape_cases}(a) and \ref{fig:ape_cases}(c), which correspond to high-variation scenarios, our model (red dashed line) aligns closely with the actual CIF curves (black solid line), capturing rapid changes and turning points more accurately than competing SOTA models. Notably, while some alternative models, such as NonStaFormer or TimesNet in Figure \ref{fig:ape_cases}(c), partially follow the trend, they tend to either smooth out peaks or lag behind sharp transitions. This limitation is particularly critical in scenarios where CIF forecasts are used for day-ahead load scheduling and operational planning. In contrast, as shown in Figures~\ref{fig:ape_cases}(b) and~\ref{fig:ape_cases}(d), the CIF series are relatively stable, and the forecasting task involves maintaining accuracy over flatter trajectories. Under these conditions, our model continues to outperform other methods, maintaining a lower error and closely tracing the observed values. These examples reinforce the effectiveness of our proposed model in adapting to both volatile and steady grid CIF patterns.

\subsection{Ablation Study}

The proposed model includes two key modules: LT-MWKC and CV-DWCC. To assess their individual impact on forecasting performance, an ablation study is conducted. The complete model is used as the baseline, and two ablation scenarios are evaluated: 1) without (w/o) LT-MWKC and 2) w/o CV-DWCC. The results in Table~\ref{tab:ablation_study} reveal insights as follows: 1) Removing LT-MWKC results in a 22.0\% increase in MAE in NSW, 42.8\% in SA, 31.8\% in QLD, and 30.3\% in VIC, highlighting the importance of localized temporal pattern extraction enriched by MFI. Without this module, the model struggles to capture short-term variations, especially under high volatility conditions like those in SA. 2) Excluding CV-DWCC results in even larger performance degradation: MAE increases by 52.5\% in NSW, 48.8\% in SA, 61.4\% in QLD, and 46.6\% in VIC. This confirms that effectively capturing dynamic cross-variable dependencies at different time-frequency resolutions is highly beneficial for accurate forecasting.

\begin{table*}[!ht]
\centering
\footnotesize
\begin{tabular}{@{}l%
>{\centering\arraybackslash}p{0.85cm}>{\centering\arraybackslash}p{0.85cm}>{\centering\arraybackslash}p{0.85cm}%
>{\centering\arraybackslash}p{0.85cm}>{\centering\arraybackslash}p{0.85cm}>{\centering\arraybackslash}p{0.85cm}%
>{\centering\arraybackslash}p{0.85cm}>{\centering\arraybackslash}p{0.85cm}>{\centering\arraybackslash}p{0.85cm}%
>{\centering\arraybackslash}p{0.85cm}>{\centering\arraybackslash}p{0.85cm}>{\centering\arraybackslash}p{0.85cm}@{}}
\toprule
\multirow{3}{*}{\textbf{Ablation setting}} 
& \multicolumn{3}{c}{\textbf{NSW}} 
& \multicolumn{3}{c}{\textbf{SA}} 
& \multicolumn{3}{c}{\textbf{QLD}} 
& \multicolumn{3}{c}{\textbf{VIC}} \\
\cmidrule(lr){2-4} \cmidrule(lr){5-7} \cmidrule(lr){8-10} \cmidrule(lr){11-13}
& \scriptsize RMSE & \scriptsize MAE & \scriptsize SMAPE 
& \scriptsize RMSE & \scriptsize MAE & \scriptsize SMAPE 
& \scriptsize RMSE & \scriptsize MAE & \scriptsize SMAPE 
& \scriptsize RMSE & \scriptsize MAE & \scriptsize SMAPE \\
\midrule
w/o LT-MWKC    
& 57.90 & 44.60 & 6.55 
& 132.89 & 100.37 & 44.59 
& 44.66 & 25.64 & 4.14 
& 101.80 & 75.46 & 9.34 \\
w/o CV-DWCC    
& 70.47 & 55.78 & 8.22 
& 137.73 & 104.56 & 46.00 
& 48.80 & 31.41 & 4.73 
& 111.54 & 84.91 & 10.47 \\
Complete model 
& \textbf{47.09} & \textbf{36.56} & \textbf{5.25} 
& \textbf{99.27} & \textbf{70.29} & \textbf{32.21} 
& \textbf{31.17} & \textbf{19.46} & \textbf{2.96} 
& \textbf{79.36} & \textbf{57.93} & \textbf{7.13} \\
\bottomrule
\end{tabular}
\captionsetup{justification=centering,singlelinecheck=false}
\caption{Ablation study results on forecasting performance in NSW, SA, QLD, and VIC.}
\captionsetup{justification=centering} 
\label{tab:ablation_study}
\end{table*}

\subsection{Interpretability}

\begin{figure}[!t]
\centering
\includegraphics[width=0.99\columnwidth]{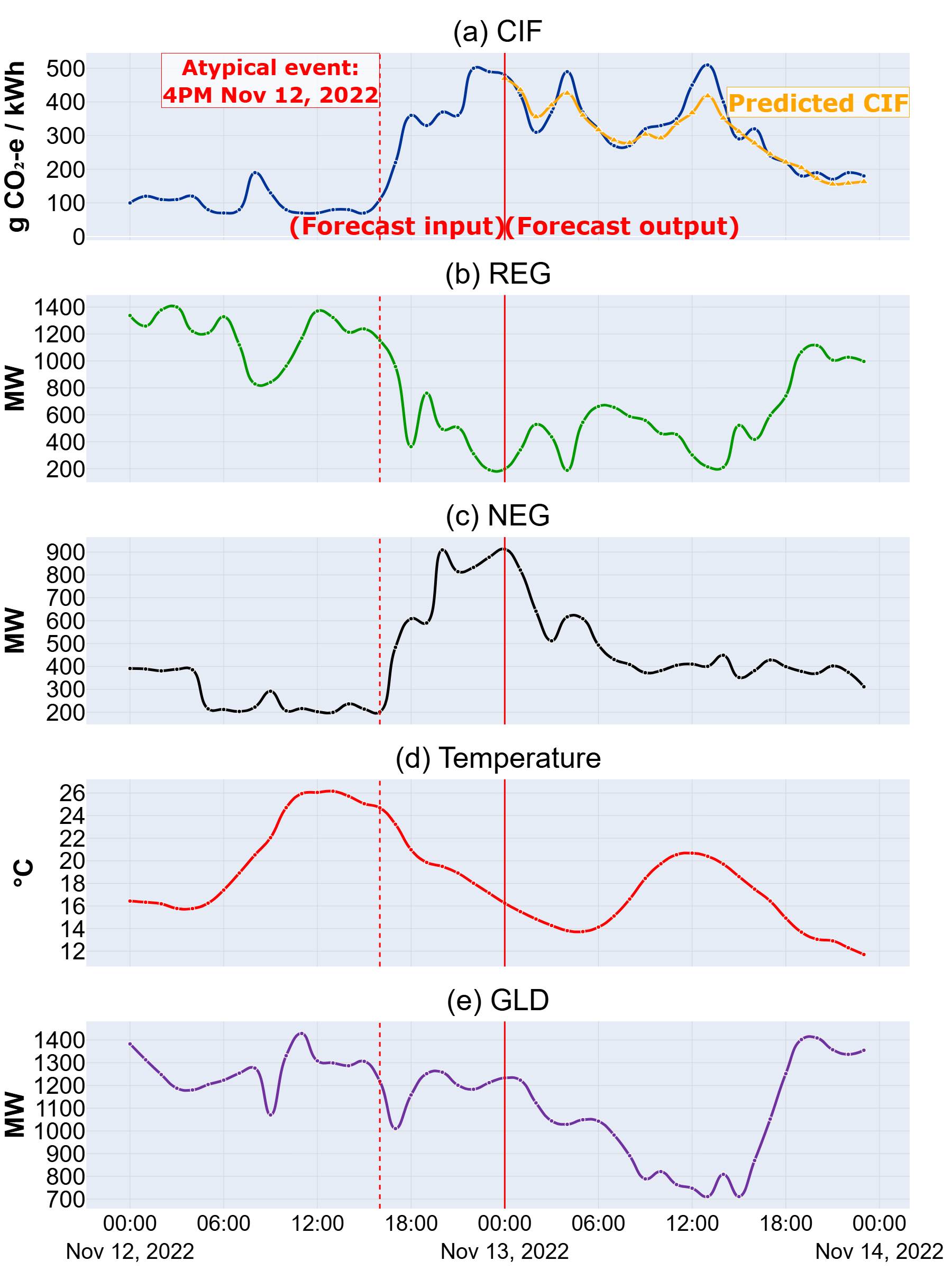}
\caption{Predicted CIF for 13 November 2022 based on MTS input from 12 November 2022.}
\label{fig:Atypical_case}
\end{figure}

Grad-CAM is an explainability technique designed for CNNs that highlights influential input regions by computing gradients of the model output relative to the convolutional feature maps. Motivated by the fact that it has been used in both computer vision \cite{F72, F73} and MTS analysis tasks \cite{F75, F74}. It is utilized to provide insights into the prediction mechanism of the proposed model.

As an exploration of model interpretability, we analyze a representative case from SA, selected for its complex CIF dynamics. At around 4:00 PM on 12 November 2022, severe weather triggered a double-circuit transmission tower failure in SA, causing both the South East–Tailem Bend 275 kV lines and the Keith–Tailem Bend 132 kV line to trip, as reported in \cite{F70, F71}. This event isolated the SA power system from the NEM, leading to frequency and voltage fluctuations. High levels of REG, particularly distributed photovoltaics, increased system management challenges during the separation. To maintain system stability, REG was curtailed, while NEG was increased to provide essential frequency control services, as shown in Figure \ref{fig:Atypical_case}. The prediction on 13 November 2022 shows good performance, indicating the ability of the model to learn post-event patterns and capture the trend following the atypical incident.

\begin{figure}[!t]
\centering

\begin{minipage}{1\columnwidth}
    \centering
    \includegraphics[width=\columnwidth]{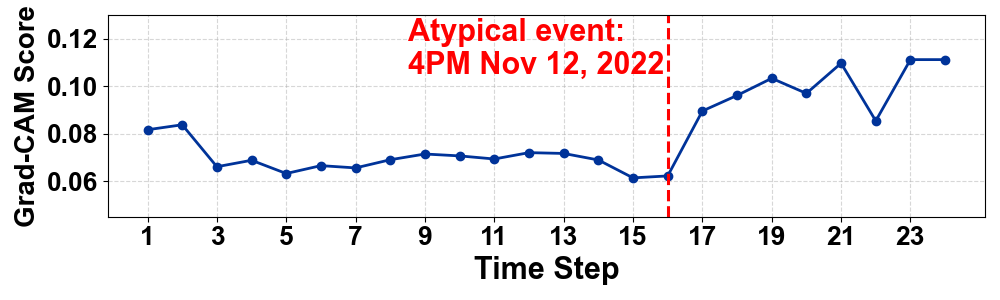}
    \textbf{(a)}
    \label{fig:Average_Grad_features}
\end{minipage}

\begin{minipage}{\columnwidth}
    \centering
    \includegraphics[width=\columnwidth]{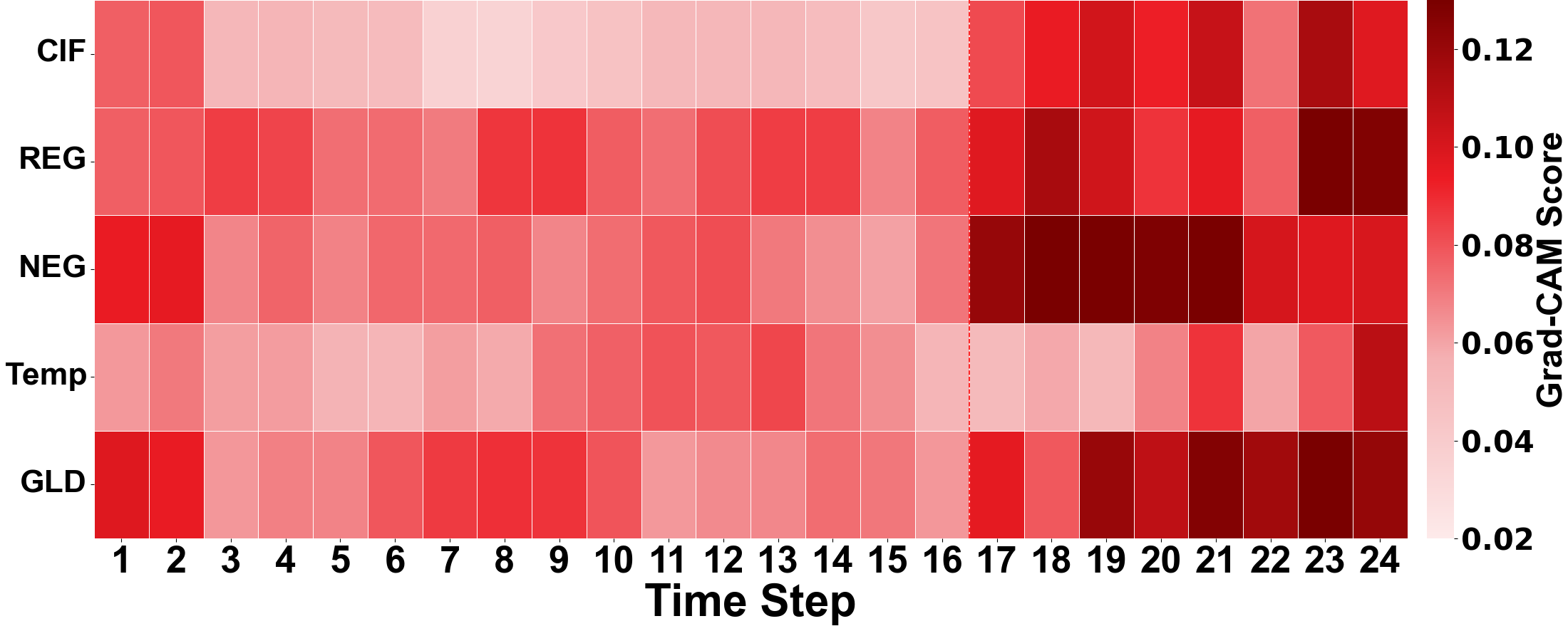}
    \textbf{(b)}
    \label{fig:Average_Grad_Scales}
\end{minipage}

\caption{Grad-CAM visualizations for 12~Nov~2022: 
(a) overall feature importance averaged across all variables, and 
(b) detailed variable-wise importance over time.}
\label{Grad_CAM_combined}
\end{figure}

Grad-CAM scores averaged across input variables and plotted over 24 time steps are shown in Figure~\ref{Grad_CAM_combined}(a). A clear rise is observed after the atypical event at time step~16 (4\,PM, November~12, 2022), indicating increased feature saliency following the outage. The importance of input variables over time steps is visualized in Figure~\ref{Grad_CAM_combined}(b), showing that during the initial period following the incident, when REG is curtailed and NEG increases, the model appropriately prioritizes NEG. Afterwards, the model shifts focus to REG as renewable output gradually stabilizes and resumes its influence on CIF. The increased feature saliency to REG may indicate that the model has learned patterns associated with severe weather that lead to large-scale curtailment, which eventually passes and results in a sudden rebound in REG. This case serves as an initial investigation into model interpretability, and future work will extend the analysis to a broader range of scenarios and regions to further validate the interpretability of the model.

\section{Conclusion}\label{Conclusion}

This study presents a novel deep learning model for short-term CIF forecasting, tailored to capture the intricate dependencies and complex patterns within the data. The proposed model integrates two parallel modules: an LT-MWKC module for capturing localized temporal patterns enriched by MFI, and a CV-DWCC module for modeling dynamic inter-variable dependencies across multiple frequencies. Empirical evaluations across electricity grids in four Australian states with different fuel mixes demonstrate that the proposed model achieves SOTA performance across all examined performance metrics. In SA, where CIF patterns are highly volatile, our proposed model shows even more pronounced gains, achieving an MAE that is 26.9\% lower than even the best-performing comparative model examined. Ablation studies confirm the complementary strengths of the LT-MWKC and CV-DWCC modules. Our analysis suggests that the model delivers interpretable insights that are practically valuable, as shown in the examination of a highly atypical grid outage event. Future work will explore the generalizability of the model across global electricity markets and focus on developing load optimization methods that actively leverage real-time CIF forecast signals.

\section*{Acknowledgments}
This study received funding from the RACE for 2030 Cooperative Research Centre, with support from Buildings Alive Pty Ltd. and the University of Technology Sydney.

\bibliography{aaai2026}

\end{document}